\documentclass[runningheads]{llncs}

\usepackage{graphicx}
\usepackage{adjustbox}
\usepackage{float}
\usepackage{tikz}
\usepackage{multirow} % Ricordati di includerlo nel preambolo!
\usepackage{pifont}
\usepackage{fontawesome5}

\usepackage{iciap}

\usepackage{iciapabbrv}

\usepackage{booktabs}

\usepackage[accsupp]{axessibility}  % Improves PDF readability for those with disabilities.

\usepackage{hyperref}

\usepackage{orcidlink}
\usepackage{ifpdf}
\ifpdf
\fi
\begin{document}

%\title{How far can we get with off-the-shelf Multimodal Large Language Model in Episodic Memory Question Answering?} 
\title{How Far Can Off-the-Shelf Multimodal Large Language Models Go in Online Episodic \\Memory Question Answering?}

\titlerunning{How Far Can Off-the-Shelf MLLMs Go in OEM-VQA?}

\author{Giuseppe Lando\textsuperscript{1,*}\orcidlink{0009-0008-4209-6852}, Rosario Forte\textsuperscript{1,*}\orcidlink{0009-0003-2334-0307},\\ Giovanni Maria Farinella\textsuperscript{1}\orcidlink{0000-0002-6034-0432}, and Antonino Furnari\textsuperscript{1}\orcidlink{0000-0001-6911-0302}}
\institute{Department of Mathematics and Computer Science, University of Catania, IT\\
\email{giuseppe.lando@studium.unict.it, rosario.forte@phd.unict.it,  \\ \{giovanni.farinella, antonino.furnari\}@unict.it}\\
\thefootnote{*Co-First Authors}}
\authorrunning{G. Lando; R. Forte; G.M. Farinella; A. Furnari.}
\maketitle
\begin{abstract}
We investigate whether \emph{off-the-shelf} Multimodal Large Language Models (MLLMs) can tackle Online Episodic-Memory Video Question Answering (OEM-VQA) without additional training.  
Our pipeline converts a streaming egocentric video into a lightweight textual memory, only a few kilobytes per minute, via an MLLM descriptor module, and answers multiple-choice questions by querying this memory with an LLM reasoner module.  
On the \textsc{QAEgo4D-Closed} benchmark, our best configuration attains \textbf{56.0\%} accuracy with \textbf{$\sim$3.6 kB per minute} storage, matching the performance of dedicated state-of-the-art systems while being $10^{4}$–$10^{5}$ times more memory-efficient.  
Extensive ablations provides insights into the role of each component and design choice, and highlight directions of improvement for future research.
\end{abstract}

\section{Introduction}
\label{sec:intro}

\begin{figure}[t]
    \centering
     \includegraphics[width=\textwidth]{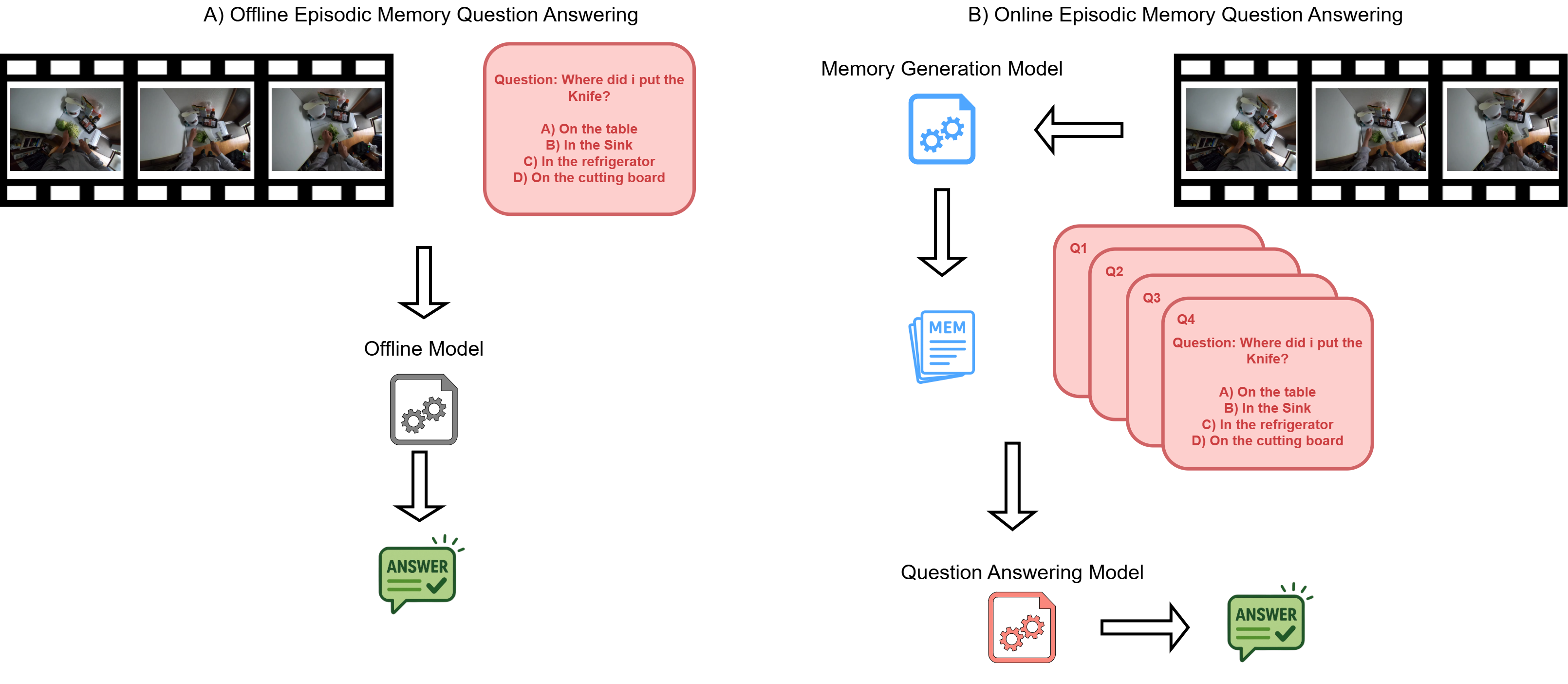}
    \caption{\textbf{Offline vs.\ online episodic-memory QA.} 
    In the \emph{offline} setting (left), the model re-processes the full video whenever it has to answer a new question. In this setting, the video itself acts as a form of high-dimensional memory and video processing happens with prior knowledge of the query.
    %every question is paired with the full video \(v\); for each new question, the model must re-process all frames before answering. Note that, in this setting, the video processing happens with prior knowledge of the query.  
    In \emph{online} settings (right), video is processed in a streaming fashion to build a compact memory, which is later referenced to answer user queries. This avoids the need to re-process the video at every frame, keeping low latency. In this case, video processing happens with no prior knowledge of the query, which represents an additional challenge.}
    
    % \(M\).  
    % The same memory \(M\) is then used to answer any number of present or future queries, eliminating repeated video processing and keeping latency per question constant. Note that in this scenario, there is no prior knowledge of the query.}
    \label{fig:overview}
\end{figure}

While performing their daily activities, humans constantly gather and organize experiences into what is known as \textit{episodic memory}~\cite{tulving1972episodic} — a powerful, flexible ability that allows us to recall relevant information with remarkable precision, even after long periods. Emulating such capabilities in artificial intelligence systems could unlock a wide range of applications, from \textit{assistive AI} to \textit{autonomous agents} operating in real-world environments.
Recently, the advent of long-form egocentric video datasets such as \textit{Ego4D}~\cite{grauman2022ego4dworld3000hours} has brought attention to the problem of \textit{episodic memory retrieval}, which is often formulated as an egocentric Video Question Answering (VideoQA) problem. In particular, the \textit{Natural Language Queries (NLQ)} task challenges models to retrieve relevant segments from long first-person videos based on free-form natural language questions. Solving this task requires not only fine-grained temporal localization but also multi-modal reasoning across long temporal horizons.
However, existing approaches consider an offline setting in which methods have access to the entire video during inference, leading to storage and computational costs that grow with video length. 
In contrast, in real application scenarios, methods should be able to process video in an online (or streaming) fashion, processing one frame at a time and storing relevant information in a compact memory, which will be referenced later to answer users' queries (see Figure~\ref{fig:overview}).

%Moreover, offline processing pipelines introduce latency and rigidity, preventing real-time interaction and flexible memory updates.

%Few works have recently tackled this issue \textcolor{red}{proposing ... }.
While current approaches to Online Episodic Memory Video Question Answering (OEM-VQA) can achieve compelling results~\cite{di2025streamingvideoquestionansweringincontext}, they are generally based on dedicated designs. 
On the other hand, current Multimodal Large Language Models (MLLMs) have shown remarkable abilities to generalize to different video understanding tasks~\cite{qwen2025qwen25technicalreport}, so it is natural to ask whether MLLMs can reasonably tackle the OEM-VQA task.
Note that, while Multimodal Large Language Models have been used for VQA tasks, OEM-VQA poses unique challenges, such as the increased length of videos (videos in Ego4D are $8.2$-minutes long in average), the ability to recall fine-grained details (e.g., Where did i put the Knife?), and the need to process the video in an online fashion.

In this paper, we investigate how far off-the-shelf MLLMS can go in OEM-VQA.
To address this question, we propose a simple yet effective pipeline made of three components: a \textit{descriptor} MLLM tasked with continuously ingesting the input video and summarize it, a \textit{memory}, made of textual descriptions produced by the descriptor model, and a \textit{reasoner} LLM, responsible for answering user queries based on the textual memory (see Figure~\ref{fig:sample_image}).
%leveraging off-the-shelf Multimodal Large Language Models (MLLMs) for online processing.
While simple, the proposed architecture follows the design principles of OEM-VQA, involving a compact memory which requires only few kilobytes of plain text per video minute, and is human-readable and interpretable.
% has three key advantages:
%     \begin{enumerate}
%     \item \textbf{Compactness and efficiency.}  
%           Storing a few kilobytes of plain text per video minute is orders of magnitude cheaper than keeping dense visual embeddings or raw frames, making our solution practical for long-horizon, always-on wearable cameras.
%     \item \textbf{Interpretability.}  
%           Textual memories are human-readable: users can inspect, edit, or filter the generated summaries without any specialised visualisation tools.
%     \item \textbf{Ease of expansion.}  
%           Because memories are plain language, they can be concatenated, re-ordered, or augmented with external knowledge (e.g. location metadata) without any re-training.  
%           The same memory can be re-prompted with entirely new questions long after the video has been processed.
%     \end{enumerate}
Experiments on Ego4D~\cite{grauman2022ego4dworld3000hours} show that our model achieves a performance on par with current OEM-VQA methods~\cite{di2025streamingvideoquestionansweringincontext, cvpr24_groundvqa} %(\textcolor{red}{56.00\% vs 56.00\%}) 
despite being much simpler and requiring less memory at inference time. 
In contrast with previous  works~\cite{cvpr24_groundvqa, di2025streamingvideoquestionansweringincontext, 9857465}, our method relies on off-the-shelf components, shedding light on the abilities of language model to tackle OEM-VQA, and providing a challenging benchmark to assess progress in MLLMs.
Towards this direction, we further show how the choice of models and parameters affects downstream performance, providing insights for future developments and research direction.
%Building on these properties, our pipeline tackles the \emph{Natural-Language Query} (NLQ) task on the \textsc{QAego4d} benchmark.
%Our study denotes the capability of MLLM to reach near SOTA performances without any specific training and in an online fashion. 
%More specifically our pipeline exploits the capability of MLLMs to produce rich and accurate video description to build a lightweight textual memory and then queries this memory with another LLM.
In sum, our contribution is two-fold: 1) A training-free pipeline for OEM-VQA, 2) A systematic ablation study investigating its optimal configuration.

%           We vary sub-clip length, decide whether or not to append the preceding temporal context, test the effect of including template questions into the caption prompt, and compare a set of lighter captioning and QA models, showing how each choice affects accuracy.
    
%     \item \textbf{A head-to-head comparison between a closed and open source MLLM} 
%               In every experiment, we compare the best-performing configuration of a closed-source Gemini model against the best-performing configuration of an open-source counterpart, LLaVA-One-Vision.
% \end{enumerate}

\section{Related Work}

\textbf{Episodic Memory VQA:}
Episodic memory retrieval was first introduced in Ego4D~\cite{grauman2022ego4dworld3000hours} as the Natural Language Query (NLQ) task: Given an egocentric video \textit{v} and a textual query \textit{q}, retrieve the temporal window \textit{t} containing the visual information relevant to answering the query. Subsequently, QAEgo4D~\cite{9857465} extended this formulation by associating an answer \textit{a} with each question \textit{q} in the original dataset, enabling the evaluation of models on open-ended answer generation. Later, in~\cite{cvpr24_groundvqa}, a closed-ended version of the task was proposed by introducing a multiple-choice paradigm for the answer \textit{a}, addressing challenges related to evaluation metrics in open-ended generation. In this setting, the goal is to infer the correct answer among four candidate options. Different methodologies have been developed to tackle the NLQ task~\cite{patel2025advancingegocentricvideoquestion}. Online methodologies have also been recently introduced. Notably,  ReKV~\cite{di2025streamingvideoquestionansweringincontext}  integrates existing MLLMs employing a key-value (KV) cache mechanism to store processed video information and enabling efficient retrieval of context relevant to a given query, which leads to state-of-the-art (SOTA) results.
In this study, we investigate the performance of off-the-shelf MLLMs to tackle the OEM-VQA when integrated in the proposed architecture.
    
\noindent
\textbf{Large Language Models (LLMs) and Video Understanding:}  
Recent advances in Video Language Models (MLLMs) have significantly impacted video‐processing tasks~\cite{li2024llava,geminiteam2024geminifamilyhighlycapable, zhang2025videollama3frontiermultimodal, wang2024internvideo2scalingfoundationmodels}. Typically, these methods use a visual encoder plus a projection layer to map visual features into the text embedding space; %the resulting visual tokens are then concatenated with textual tokens and fed into an LLM to generate the output. 
However, they often rely on token‐compression techniques to keep sequence lengths manageable, which can incur substantial information loss when dealing with longer videos. Consequently, they are not well suited for long‐range or online video understanding.
In this work, we exploit these models’ strength by generating textual captions on short‐context video every \textit{s} seconds. This strategy allows us to handle long video streams without altering the underlying VLLM architecture.

\noindent
\textbf{Online and Long Video Processing:}
To process long egocentric streams under low resource budgets, many systems employ sliding-window attention or token compression to limit the length of token sequences, but at the cost of discarding temporal dependencies~\cite{NEURIPS2021_6a30e32e,di2025streamingvideoquestionansweringincontext}.
Alternative approaches maintain a key–value (KV) cache or distilled motion summaries of past frames to enable efficient lookup of relevant context. 
%ReKV~\cite{di2025streamingvideoquestionansweringincontext}, for instance, stores encoded video KV-caches off-device and retrieves only the entries most similar to the incoming query.
%Such memory modules allow selective retrieval of pertinent visual context with minimal latency, making online processing of egocentric videos feasible. However, they often lack explainability and do not fully exploit the information extracted from the video. 

Particularly relevant to our work, the authors of~\cite{yang2025egolifeegocentriclifeassistant} propose EgoButler, an integrated system combining EgoGPT, an omni-modal model trained on egocentric datasets, and EgoRAG, a retrieval-based component supporting ultra-long-context question answering.
While our approach is similar in spirit, EgoButler tackles offline long-range VQA, while we explicitly focus on online episodic memory question answering.
Overall, our goal is to investigate how off-the-shelf MLLMs can support OEM-VQA.
% Unlike their approach, which relies on retrieval-augmented generation (RAG) methods, we propose to build a textual memory that describes the input video in an online fashion. As the video is processed, this memory is incrementally built and updated. When a user query is issued, the accumulated memory is passed along with the query to a reasoner model to produce the answer.

\section{Method}
\begin{figure}[t] 
    \centering
     \includegraphics[width=\textwidth]{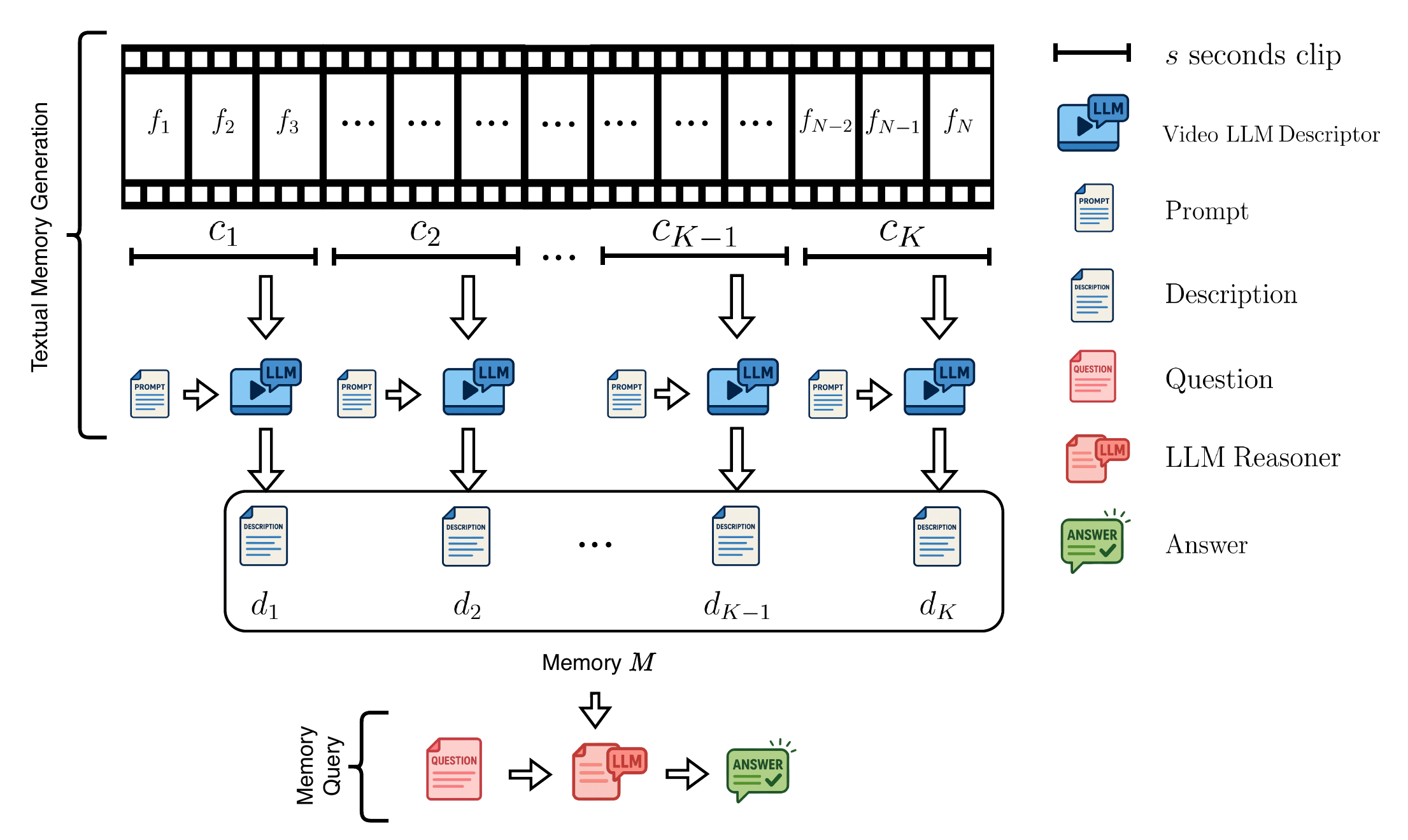}
    \caption{\textbf{Method Overview} The input egocentric video is divided into non-overlapping clips $(c_1, c_2, \dots, c_K)$, each processed independently by a Video LLM Descriptor to generate a textual description $d_k$. These descriptions are concatenated to form a compact textual memory $M$, which is persistent and independent. At query time, a natural language question is combined with the accumulated memory and passed to an LLM Reasoner, which produces the final answer. The system operates in a streaming fashion, enabling online memory update at query time.}
    \label{fig:sample_image}
\end{figure}

Our method (Figure~\ref{fig:sample_image}) achieves OEM-VQA in two main stages: 1) textual memory generation, 2) memory query, which are described in the following. %The following sections describe these two stages.
%We design a two-stage online pipeline that constructs a compact textual memory from streaming egocentric video and leverages it for close-ended question answering. In the following sections, we describe the memory construction and query answering stages in detail.

\subsection{Textual Memory Generation}

Given an incoming egocentric video stream $v$, the goal of this stage is to construct a lightweight, human-readable memory $M$ that summarizes the content of $v$ without retaining raw visual data (see Figure~\ref{fig:sample_image}-Textual Memory Generation). Formally, let $v = (f_1, f_2, \dots, f_N)$ be a sequence of $N$ frames, where each frame $f_i \in \mathcal{F} = [0, 1]^{W \times H \times 3}$ represents an RGB image.
We segment the video $v$ into a sequence of non-overlapping clips $(c_1, c_2, \dots, c_K)$, where each clip $c_k$ contains a contiguous sequence of frames spanning a fixed temporal duration $s$ seconds. The clip duration $s$ is a configurable parameter, and different values are explored in our experiments to assess the trade-offs between memory granularity and overall performance.
For each clip $c_k$, we use a descriptor model (which is a MLLM) to generate a textual description $d_k$. To guide the generation, we provide the model with a structured prompt that instructs it to describe the scene from a first-person perspective.
The memory $M$ is defined as the concatenation of the clip-level descriptions:
\begin{equation}
M = (d_1, d_2, \dots, d_K).    
\end{equation}
Optionally, when constructing the prompt for the captioning model, we include the description $d_{k-1}$ of the previous clip $c_{k-1}$ to provide previous context for $c_k$. %This mechanism allows the descriptor to retain a limited sense of temporal coherence across consecutive clips, at the cost of increased prompt length.
This design enables online memory construction, as each $d_k$ can be generated immediately after observing $c_k$, without the need for storing the full video $v$.

\subsubsection{Prompt Design}

\begin{figure}[t]
    \centering
    \includegraphics[width=\textwidth]{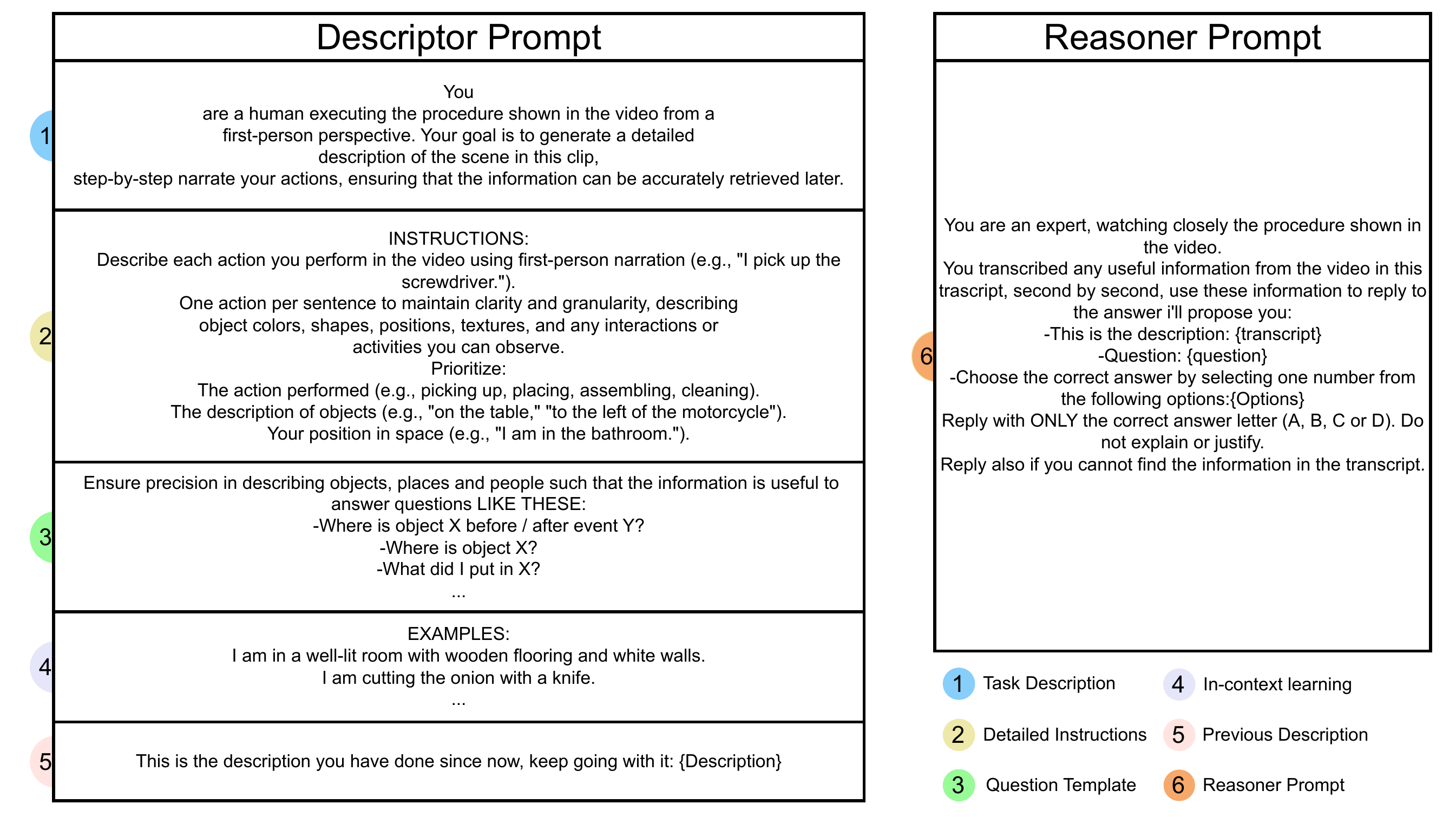}
    \caption{Our Descriptor prompt (left) follows a rigid structure consisting of five main sections: 
    {\textcolor[HTML]{87CEFA}{$\bullet$}} Task description; 
    {\textcolor[HTML]{EEE8AA}{$\bullet$}} Detailed instructions; 
    {\textcolor[HTML]{98FB98}{$\bullet$}} Question templates; \newline
    {\textcolor[HTML]{E6E6FA}{$\bullet$}} In-context examples;
    and 
    {\textcolor[HTML]{FFE4E1}{$\bullet$}} Previous descriptions.
    {\textcolor[HTML]{F6A96B}{$\bullet$}} Reasoner prompt example (right). }
    \label{fig:Prompt}
\end{figure}

The prompt provided to the captioning model is designed to obtain a detailed description of the scene, including information on the observed objects and the location of the camera wearer. With this purpose, we created a structured prompt, as reported in Figure~\ref{fig:Prompt}-left. The prompt comprises five core sections: 1) Task Description: instructs the MLLM on the general goal the its task; 2) Detailed Instruction: provides details on the task and expected output; 3) Question Template: includes templates of questions that may be answered later, to encourage the inclusion of useful details. In our experiments, we consider the templates provided to the annotators of NLQ in Ego4D~\cite{grauman2022ego4dworld3000hours}; 4) In-context Learning: provides an output example to clarify what the expected output is; 5) Description Continuity: reports the description generated at the previous step by the model to ensure continuity and add more context on the scene. For the reasoner we used a standard prompt Figure~\ref{fig:Prompt}-Right, giving the content of the memory and a brief introduction.
For our experiments, we also tested different prompt configurations by selectively removing the Question Template section and/or the Previous Description section to evaluate the impact of these components on the model’s ability to produce accurate and complete descriptions.

\subsection{Memory Query}

%Once the textual memory $M$ has been constructed, our goal is to answer natural language queries without accessing the original video frames. 
Given a query $q$, we use \textit{reasoner} module(which is an LLM) to read the memory $M$ and select the most appropriate answer among a set of candidate choices (Figure~\ref{fig:overview} Memory Query).
In the Close QA setting that we adopt, each question $q$ is associated with four candidate answers. The model is tasked with selecting the correct one, rather than generating a free-form text response. Therefore, the output $a$ belongs to the set $\{A, B, C, D\}$, where each letter corresponds to one of the candidate options.
The \textit{reasoner} LLM receives the entire memory $M$, the query $q$, and the list of candidate answers concatenated into a single prompt. 
%A short preamble is included in the prompt to specify the task objective and to encourage expert-level reasoning over the provided memory.
%
No retrieval, filtering, or re-encoding of the visual stream is performed at this stage, since the model operates purely over the textual domain.
Formally, the answer $a$ is produced by the reasoner module $Q$ as $a = Q(m, q),$ where $a \in \{A, B, C, D\}$. The prompt used in this stage is reported in Figure~\ref{fig:Prompt}-right.

% \subsection{Models}
% We employ two categories of models in our pipeline: captioning models to construct the textual memory, and question-answering (QA) models to select answers based on the memory and the query.
% For the scene description stage, we explore two main families of Multimodal Large Language Models (MLLMs). From the Gemini 2.0 family, we consider \texttt{Gemini 2.0 Flash}, \texttt{Gemini 2.0 Flash Lite}, and the experimental \texttt{Gemini 2.0 Flash Thinking} version (\texttt{gemini-2.0-flash-thinking-exp-01-21}), which introduces a lightweight reasoning step before producing outputs. As an open-source counterpart, we adopt LLaVA-One-Vision models, specifically \texttt{LLaVA-One-Vision 7B} and its smaller variant \texttt{0.5B}, which leverage the Qwen-2 language backbone to strengthen visual-textual alignment.

% For question answering over the textual memory, we primarily rely on \texttt{DeepSeek R1}, a reasoning-enhanced variant derived from \texttt{DeepSeek V3}. While DeepSeek V3 serves as the base language model, R1 introduces explicit reasoning mechanisms to improve performance on complex multi-step queries. In addition to DeepSeek models, we also conduct ablation studies using other open-source language models, including \texttt{Qwen 2.5 70B} and \texttt{LLaMA 3.1 405B}.

\section{Experimental Settings and Results}

We evaluate the proposed appraoch on the QAEgo4D-Closed~\cite{cvpr24_groundvqa} benchmark, a variant of  Ego4D NLQ task in which models are tasked with selecting the correct answer among four candidate options. Accuracy is used as the primary evaluation metric across all experiments.

\subsection{Comparison with the state of the art}
We start by evaluating the proposed approach with previous state-of-the-art methods. For these comparisons we consider two configurations of our model. The first one is based on the closed-source \texttt{Gemini 2.0 Flash Thinking} used to provide video descriptions every $30s$ of the observed video. The second one is based on the open-source \texttt{LLaVaOneVision 7B}, which generate descriptions every $15s$. Both models use \texttt{DeepSeek R1} as a reasoner. Additional details on these configurations and ablations of model choices are provided in the next sections.
As compared methods, we consider ReKV LLaVaOneVision 7B and 0.5B~\cite{di2025streamingvideoquestionansweringincontext}, as well as GroundVQA baseline~\cite{cvpr24_groundvqa}, which are, to our knowledge, the only state-of-the-art approaches tackling the considered task published to date.
We report results in Table~\ref{tab:methods_comparison}, including for each method the size of the auxiliary memory used to answer queries at test time and the model accuracy.
%As a first step, we compare the performance of our best configurations against existing methods, reporting memory requirements and accuracy. 
%\textcolor{red}{We consider the following configurations (ablations are provided in the following sections) ...}
%\textcolor{red}{The proposed method is compared against the following state-of-the-art online VQA methods ...}
%This comparison highlights that our method achieves comparable accuracy to state-of-the-art models while significantly reducing memory footprint. 
As can be seen, our method allows to obtain performance on par with current methods, but with great advantages in terms of memory consumption. 
%Among competitors, only Ground VQA produces indeed 
%We compare our best-performing configurations with existing state-of-the-art (SOTA) methods in terms of memory efficiency and task accuracy, as summarized in Table~\ref{tab:methods_comparison}. 
Indeed, leveraging off-the-shelf MLLMs to build a textual memory significantly reduces memory requirements while maintaining competitive performance.
Specifically, our Gemini-based architectures achieves an accuracy of 56.00\%, and LLaVaOneVision-based architectures reach 51.88\%, closely matching the SOTA results achieved by RekV-LLaVaOneVision 7B (56\%). Importantly, our models achieve these results without any fine-tuning and relying on off-the-shelf components, highlighting the strength of the zero-shot capabilities of modern MLLMs.
While performance is on-par with current approaches, our method obtains ubstantial gains in terms of memory usage, requiring only 3.57 KB per minute of video (Gemini-based) and 5.05 KB per minute (LLaVaOneVision-based). 
This brings improvements of several orders of magnitude compared to RekV-LLaVaOneVisionv 0.5B (68.92 MB/minute) and RekV-LLava1v7 (329.64 MB/minute).
Notably, our models also outperform Ground VQA both in terms of memory and accuracy, which uses only 1.29 MB per minute, while achieving a reduced accuracy of 48.7\% accuracy, outperformed by both our configurations.
%Overall, these findings underscore the effectiveness of our pipeline, which combines off-the-shelf models without the need for task-specific fine-tuning or heavy infrastructure, offering a scalable and practical solution for episodic memory question answering.

\begin{table}[t]
\caption{Comparison with the state of the art in terms of memory usage (KB or MB per minute of processed video) and accuracy on the QAego4D-Closed benchmark.}
\label{tab:methods_comparison}
\centering
\setlength{\tabcolsep}{8pt}
\resizebox{0.8\textwidth}{!}{%
\begin{tabular}{lcc}
\toprule
\textbf{Method} & \textbf{Memory}$\downarrow$ & \textbf{Accuracy(\%)}$\uparrow$ \\ \midrule
Ground VQA~\cite{cvpr24_groundvqa} & 1.29 MB/min & 48.70\%\\ %\hline
RekV-LLava1v 0.5~\cite{di2025streamingvideoquestionansweringincontext} & 68.92 MB/min & 50.00\%\\ %\hline
RekV-LLaVaOneVision 7B~\cite{di2025streamingvideoquestionansweringincontext} & 329.64 MB/min & \textbf{56.00}\%\\ \midrule
LLaVaOneVision-based (Ours) & 5.05 KB/min & 51.88\%\\
Gemini-based (Ours) & \textbf{3.57 KB/min} & \textbf{56.00} \%\\ \bottomrule
\end{tabular}%
}
\end{table}

\subsection{Ablation Studies}
In this section, we report ablations on the choice of MLLM as descriptor and the description granularity (size of video segments $c_i$), the effect of including context (the previous description) in the prompt, the effect of including question templates in the prompt, and the effect of using different LLMs as reasoners. 

\subsubsection{Choice of MLLM as Descriptor and Clip Length}
Table~\ref{tab:ablation_results} compares the two considered MLLMs (\texttt{Gemini Flash Thinking} and \texttt{LLaVaOneVision 7B}) with three different clip lengths $s\!\in\!\{30,15,5\}$\,. The prompt used for these experiments keeps \emph{both} previous-clip context and template questions. We use \texttt{DeepSeek R1} as a reasoner for these experiments. 
For $s\!=\!15$ and $5$s the context for the QA model gets too large to fit in the reasoner context window, hence, at query time, we only keep one caption every 30s.
%to fit the QA token limit, an admittedly coarse but practical workaround, results at \textcolor{red}{Table }\ref{tab:ablation_results}.
\texttt{Gemini Flash Thinking} peaks at 51.6\% with $30s$ clips, while shorter segments give no benefit in this configuration.  
\texttt{LLaVA-One-Vision 7B} gets its best result at $15s$ (47. 9\%), suggesting that shorter videos are easier to describe more accurately for this model.
A context window of $5s$ leads to reduced performance in both models, with the difference being larger in the case of \texttt{LLaVA-One-Vision 7B}. This suggests that too detailed descriptions are not necessary to improve performance and that \texttt{Gemini Flash Thinking} is less sensitive to video clip length.
Given these results, we use a clip length of $30s$ for Gemini and $15s$ for LLaVaOneVision in all subsequent experiments.
%Both models drop their performance at 5s, showing that aggressive subsampling wastes the extra detail.
%Granularity must balance caption density and prompt length; a smarter retrieval scheme could unlock the gains of finer segmentation without overrunning the LLM context window.

\begin{table}[t]
\caption{Ablation study of different descriptor models and clip lengths.}
\label{tab:ablation_results}
\setlength{\tabcolsep}{8pt}
\centering
\resizebox{\textwidth}{!}{%
\begin{tabular}{cccccc}
\toprule
\textbf{Descriptor} & \textbf{Clip Length} & \textbf{Context} & \textbf{Template} &
\textbf{Reasoner} & \textbf{Accuracy(\%)}$\uparrow$\\
\midrule
\multirow{3}{*}{\parbox{3cm}{\centering Gemini 2.0\\Flash Thinking}}  % col 1
  & 30s & Yes & Yes &
    \multirow{3}{*}{DeepSeek R1} % col 5
    & \textbf{51.60\%}\\
  & 15s & Yes & Yes & & 50.00\%\\
  & 5s  & Yes & Yes & & 51.20\%\\
\midrule
\multirow{3}{*}{\parbox{3cm}{\centering LLaVaOneVision 7B}}           % col 1
  & 30s & Yes & Yes &
    \multirow{3}{*}{DeepSeek R1} % col 5
    & 37.00\%\\
  & 15s & Yes & Yes & & \textbf{47.92\%}\\
  & 5s  & Yes & Yes & & 45.74\%\\
\bottomrule
\end{tabular}}%
\end{table}

\subsubsection{Previous Caption as Context}
Table~\ref{tab:model_context_comparison} compares methods when the description generated for the previously observed clip is provided as context (see Figure~\ref{fig:Prompt}-left).
%Using the best clip granularity found per family, we evaluate the impact of including or omitting the caption of the previous clip during memory construction.
%
\begin{table}[t]
\caption{Effect of using the previous clip description as context.}
\label{tab:model_context_comparison}
\setlength{\tabcolsep}{8pt}
\centering
\resizebox{\textwidth}{!}{%
\begin{tabular}{c c c c c c}
\toprule
\textbf{Descriptor} & \textbf{Clip Length} & \textbf{Context} &
\textbf{Template} & \textbf{Reasoner} & \textbf{Accuracy(\%)}$\uparrow$\\
\midrule
\multirow{2}{*}{\parbox{3cm}{\centering Gemini 2.0\\Flash Thinking}}
  & 30s & Yes & Yes &
    \multirow{2}{*}{DeepSeek R1} & 51.60\%\\
  & 30s & No  & Yes & & \textbf{56.00\%}\\
\midrule
\multirow{2}{*}{\parbox{3cm}{\centering LLaVaOneVision 7B}}
  & 15s & Yes & Yes &
    \multirow{2}{*}{DeepSeek R1} & 47.92\% \\
  & 15s & No  & Yes & & \textbf{51.88\%} \\
\bottomrule
\end{tabular}}
\end{table}
As can be noted, removing the previous caption (Context=No) lifts accuracy from \textbf{51.6\%} to \textbf{56.0\%} for \texttt{Gemini-Flash-Thinking} and
from \textbf{47.9\%} to \textbf{51.9\%} for \texttt{LLaVA-One-Vision 7B}.  
This result suggests that including the previous context gives rise to \emph{hallucination
snowballing}~\cite{zhang2024how}: once a caption already containing inaccuracies, is appended to the next prompt, the model is pushed to stay coherent with those imperfections, allowing the error to propagate. 
Eliminating the previous caption as context removes that noisy seed and shortens the prompt, reducing contextual entropy, and freeing attention for the new visual input.
Given this observation, we disable context in the next experiments.

\subsubsection{Use of Question Templates}
%Starting from the best granularity and context configuration for each family, 
Table~\ref{tab:model_prompt_comparison} shows the impact of injecting the \emph{template-question} block into the captioning prompt (see Figure~\ref{fig:Prompt}-left).  
Adding the template yields a consistent boost for both model families.  
With \texttt{Gemini 2.0 Flash Thinking}, accuracy rises from \textbf{50.8\%} to \textbf{56.0\%}, whereas with \texttt{LLaVA-One-Vision 7B} the improvement is dramatic: from \textbf{32.2\%} to \textbf{51.9\%}.
\begin{table}[t]
\caption{Effect of including question templates in the prompt.}
\label{tab:model_prompt_comparison}
\setlength{\tabcolsep}{8pt}
\centering
\resizebox{\textwidth}{!}{%
\begin{tabular}{c c c c c c}
\toprule
\textbf{Descriptor} & \textbf{Clip Length} & \textbf{Context} &
\textbf{Template} & \textbf{Reasoner} & \textbf{Accuracy(\%)}$\uparrow$\\
\midrule
\multirow{2}{*}{\parbox{3cm}{\centering Gemini 2.0\\Flash Thinking}}
  & 30s & No & Yes    &
    \multirow{2}{*}{DeepSeek R1} & \textbf{56.00\%}\\
  & 30s & No & No & & 50.80\%\\
\midrule
\multirow{2}{*}{\parbox{3cm}{\centering LLaVaOneVision 7B}}
  & 15s & No & Yes    &
    \multirow{2}{*}{DeepSeek R1} & \textbf{51.88\%} \\
  & 15s & No & No & & 32.16\% \\
\bottomrule
\end{tabular}}
\end{table}
These improvements suggest that an explicit list of question patterns acts as \emph{soft supervision}, directing the descriptor module to store information that will be later useful to the downstream reasoner. We include template questions in our next experiments.
%Nonetheless, the absolute best results for \emph{both} families are obtained with the template section, underlining that prompt engineering is a critical lever for this task, especially when no fine-tuning is performed.

\subsubsection{Choice of LLM as Reasoner}
Table~\ref{tab:ablation_study_models} compares different LLMs as reasoner: (\texttt{DeepSeek R1}, \texttt{DeepSeek V3}, \texttt{Qwen 2.5 70B}, \texttt{LLaMA 3.1 405B}).
\begin{table}[t]
\caption{Ablation study of different reasoner LLM models.}
\label{tab:ablation_study_models}
\setlength{\tabcolsep}{8pt}
\centering
\resizebox{\textwidth}{!}{%
\begin{tabular}{c c c c c c}
\toprule
\textbf{Descriptor} & \textbf{Clip Length} & \textbf{Context} &
\textbf{Template} & \textbf{Reasoner} & \textbf{Accuracy(\%)}$\uparrow$ \\
\midrule
\multirow{4}{*}{\parbox{3cm}{\centering Gemini 2.0\\Flash Thinking}}
  & 30s & No & Yes & Qwen 2.5 70B  & 52.60\%\\
  & 30s & No & Yes & DeepSeek R1   & \textbf{56.00\%}\\
  & 30s & No & Yes & DeepSeek V3   & 51.80\%\\
  & 30s & No & Yes & LLaMa 3.1 405B & 49.40\%\\
\midrule
\multirow{4}{*}{\parbox{3cm}{\centering LLaVaOneVision 7B}}
  & 15s & No & Yes & Qwen 2.5 70B   & 50.11\% \\
  & 15s & No & Yes & DeepSeek R1    & \textbf{51.88\%} \\
  & 15s & No & Yes & DeepSeek V3    & 47.69\% \\
  & 15s & No & Yes & LLaMa 3.1 405B & 43.96\% \\
\bottomrule
\end{tabular}}
\end{table}
As shown in the table, the performance of the overall pipeline remains stable across different LLMs, highlighting the robustness of the memory representation. Models with stronger reasoning capabilities, such as \texttt{DeepSeek R1} and \texttt{Qwen 2.5 70B}, consistently achieve higher accuracy compared to other alternatives. 
Differently from \texttt{DeepSeek R1}, \texttt{DeepSeek V3} does not include Chain-of-Thought reasoning, which allows to boost performance ($56.0\%$ vs $51.8\%$ in the case of Gemini and $51.88\%$ vs $57.69\%$ in the case of LLaVaOneVision).
Among all tested models, \texttt{DeepSeek R1} yields the best overall results across both Gemini and LLaVaOneVision backbones.

\subsection{Computational Optimization}
Our best performing configuration makes use of \texttt{Gemini 2.0 Flash Thinking} as a descriptor and \texttt{DeepSeek R1} as a reasoner. While this configuration is training-free and achieves good performance, the two considered models are demanding in terms of computation.
In this section, we explore additional configurations based on lightweight descriptor and reasoner models to evaluate trade-offs between task performance and computational efficiency. All experiments keep the best setting per model family. 
As shown in Table~\ref{tab:ablation_computational_optimization}, descriptor models from the Gemini family maintain stable performance even when moving to lighter variants, with only minor fluctuations. 
Indeed, using \texttt{Gemini 2.0 Flash Lite} only brings minor decrements in accuracy ($56.00\%$ vs $55.60\%$).
In contrast, for the LLaVaOneVision family, a significant performance drop is observed when moving from the 7B version to the 0.5B version, resulting in a decrease of approximately 10\% in accuracy ($51.88\%$ vs $41.54\%$).
%We also assess the impact of using a smaller LLM, specifically Qwen 2.5 7B. %%Qua ne metto 2 tabelle

\begin{table}[t]
\caption{Ablation of lightweight descriptor models.}
\label{tab:ablation_computational_optimization}
\setlength{\tabcolsep}{8pt}
\centering
\resizebox{0.95\textwidth}{!}{%
\begin{tabular}{c c c c c c}
\toprule
\textbf{Descriptor} & \textbf{Clip Length} & \textbf{Context} &
\textbf{Template} & \textbf{Reasoner} & \textbf{Accuracy(\%)}$\uparrow$\\
\midrule
Gemini 2.0 Flash Thinking & 30s & No & Yes &
  DeepSeek R1 & \textbf{56.00}\%\\
Gemini 2.0 Flash          & 30s & No & Yes & DeepSeek R1 & 55.40\%\\
Gemini 2.0 Flash Lite     & 30s & No & Yes & DeepSeek R1 & 55.60\%\\
\midrule
LLaVaOneVision 7B         & 15s  & No  & Yes  &
  DeepSeek R1  & \textbf{51.88}\%\\
LLaVaOneVision 0.5B       & 15s  & No  & Yes  & DeepSeek R1 & 41.54\%\\
\bottomrule
\end{tabular}}
\end{table}
Table~\ref{tab:ablation_qa_optimization} shows the impact of using the smaller reasoner model Qwen 7B instead of DeepSeek R1, which leads to a notable decrease in performance across both lightweight pipelines, confirming the critical role of the reasoning module in sustaining overall task accuracy.
Nevertheless, results suggest that reasonable performance can still be achieved using lightweight descriptor and reasoner models. 

\begin{table}[t]
\caption{Ablation of lightweight reasoner models.}
\label{tab:ablation_qa_optimization}
\setlength{\tabcolsep}{8pt}
\centering
\resizebox{\textwidth}{!}{%
\begin{tabular}{c c c c c c}
\toprule
\textbf{Descriptor} & \textbf{Clip Length} & \textbf{Context} &
\textbf{Template} & \textbf{Reasoner} & \textbf{Accuracy(\%)}$\uparrow$\\
\midrule
\multirow{2}{*}{\parbox{3cm}{\centering Gemini 2.0\\Flash Lite}}
  & 30s & No & Yes &  DeepSeek R1 & \textbf{55.60}\%\\
  & 30s & No & Yes &  Qwen 7B        & 46.60\%\\
\midrule
\multirow{2}{*}{\parbox{3cm}{\centering LLaVaOneVision 0.5B}}
  & 15s& No& Yes& DeepSeek R1& \textbf{41.54}\%\\
  & 15s& No& Yes& Qwen 7B& 34.07\%\\
\bottomrule
\end{tabular}}
\end{table}

\subsection{Qualitative Examples}

\begin{figure}[t] 
    \centering
     \includegraphics[width=\textwidth]{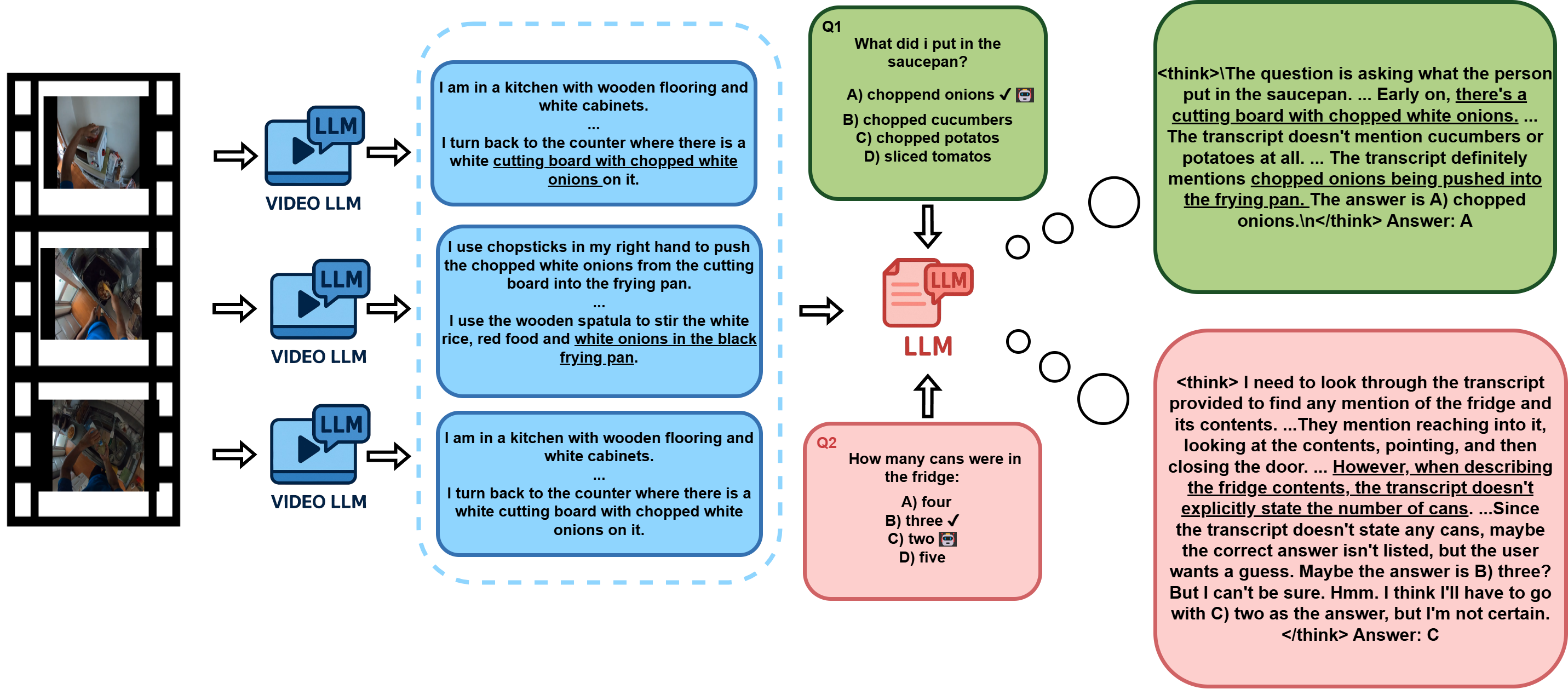}
      \caption{Qualitative examples from the validation set.  
           \textbf{Left:} Video clips are processed by the descriptor into textual memory entries.  
           \textbf{Centre:} two multiple-choice questions are presented to the reasoner, with a \ding{52} on the correct answer and a \faRobot~on the answer chosen by the reasoner.  
           \textbf{Right:} the model’s chain-of-thought reveals how it searches the memory to identify supporting evidence.  
           In the top example, the relevant fact is clearly present in the memory, enabling a correct answer.
           In the bottom example, the information is missing, leading the model to speculate and fail.}
    \label{fig:qual-example}
\end{figure}

Figure~\ref{fig:qual-example} illustrates two qualitative examples showing how the reasoner model reasons over the textual memory.  
When relevant information is explicitly present in the captions, the model retrieves it effectively and answers correctly (top example, Q1).  
Conversely, when key details are missing or ambiguously described, the model tends to speculate, leading to errors (bottom example, Q2).  
This highlights how answer quality is tightly linked to the completeness and precision of the generated memory.

\section{Conclusion and Outlook}
This study explored the potential of \emph{off-the-shelf} Multimodal Large Language Models (MLLMs) for Online Episodic-Memory Video Question Answering (OEM-VQA).  
We proposed a training-free two-stage pipeline that first transforms an egocentric video stream into a compact, human-readable textual memory and then reasons over this memory with a language-only LLM.
Experiments on \textsc{QAEgo4D-Closed} show that, without any task-specific fine-tuning, the Gemini-based configuration attains a state-of-the-art accuracy of \textbf{56.0\,\%}, while requiring only \textbf{$\sim$3.6 kB per minute} of memory, several orders of magnitude lighter than prior methods.
Ablations illustrated how model choices such as including template questions and choosing the right clip length can greatly affect performance, while lightweights configurations are also attainable, provided that some accuracy is sacrificed.
%Careful prompt design (first-person narration, question templates) emerged as a stronger lever than model size, both improving accuracy.
%\textbf{Limitations.} Because the video is discarded after captioning, any hallucinations or omissions in the clip descriptions propagate uncorrected; moreover, reasoning over very long memories is still bounded by current context windows.
% While we show the potential of MLLMs to address OEM-VQA, our analysis also highlights three main directions of improvement. \textbf{Memory construction} is sensitive to clip length, suggesting that it can benefit from adaptive clip segmentation strategies to keep accurate descriptions without inflating storage and overflowing context length. 
% The \textbf{Descriptor} is sensitive to prompt, benefiting from query-aware templates and being hindered from the inclusion of past descriptions as context. This suggests that, while we can use MLLMs in a training-free fashion, a careful prompt design or a task-aware fine-tuning may be key to performance.
% Beyond model size, \textbf{Reasoner} performance is affected by inference strategy, with techniques such as chain-of-thought leading to better results. Also, the reasoner struggles with managing longer contexts arising from shorter clip lengths, which may be mitigated by the inclusion of techniques able to select sub-sections of the memory relevant to the current query.
% We hope that our investigation will encourage more research in these directions.
While showcasing the potential of MLLMs for OEM-VQA, our analysis reveals three key areas for improvement. \textbf{Memory construction} is sensitive to clip length, suggesting benefits from adaptive segmentation to retain accuracy without exceeding storage or context limits. The \textbf{Descriptor} is prompt-sensitive, benefiting from query-aware templates but hindered by including prior descriptions, indicating that careful prompt design or task-aware fine-tuning may be crucial. \textbf{Reasoner} performance, beyond model size, depends on inference strategy (e.g., chain-of-thought), and struggles with long contexts from short clips; this could be addressed via relevance-based memory selection. We hope our findings inspire further research in these directions.

\section{Acknowledgment}
This research has been funded by the European Union - Next Generation EU, Mission 4 Component 1 CUP E53D23016240001 - Project PRIN 2022 PNRR TEAM and FAIR – PNRR MUR Cod. PE0000013 - CUP: E63C22001940006.

\par\vfill\par

\bibliographystyle{splncs04}
\bibliography{main}

\begin{thebibliography}{10}
\providecommand{\url}[1]{\texttt{#1}}
\providecommand{\urlprefix}{URL }
\providecommand{\doi}[1]{https://doi.org/#1}

\bibitem{9857465}
Bärmann, L., Waibel, A.: Where did i leave my keys? — episodic-memory-based question answering on egocentric videos. In: 2022 IEEE/CVF Conference on Computer Vision and Pattern Recognition Workshops (CVPRW). pp. 1559--1567 (2022). \doi{10.1109/CVPRW56347.2022.00162}

\bibitem{cvpr24_groundvqa}
Di, S., Xie, W.: Grounded question-answering in long egocentric videos. In: CVPR (2024)

\bibitem{di2025streamingvideoquestionansweringincontext}
Di, S., Yu, Z., Zhang, G., Li, H., Zhong, T., Cheng, H., Li, B., He, W., Shu, F., Jiang, H.: Streaming video question-answering with in-context video kv-cache retrieval (2025), \url{https://arxiv.org/abs/2503.00540}

\bibitem{grauman2022ego4dworld3000hours}
Grauman, K., et~al.: Ego4d: Around the world in 3,000 hours of egocentric video (2022), \url{https://arxiv.org/abs/2110.07058}

\bibitem{li2024llava}
Li, B., Zhang, Y., Guo, D., Zhang, R., Li, F., Zhang, H., Zhang, K., Li, Y., Liu, Z., Li, C.: Llava-onevision: Easy visual task transfer. arXiv preprint arXiv:2408.03326  (2024)

\bibitem{patel2025advancingegocentricvideoquestion}
Patel, A., Chitalia, V., Yang, Y.: Advancing egocentric video question answering with multimodal large language models (2025), \url{https://arxiv.org/abs/2504.04550}

\bibitem{qwen2025qwen25technicalreport}
Qwen, :, et~al.: Qwen2.5 technical report (2025), \url{https://arxiv.org/abs/2412.15115}

\bibitem{NEURIPS2021_6a30e32e}
Ryoo, M., Piergiovanni, A., Arnab, A., Dehghani, M., Angelova, A.: Tokenlearner: Adaptive space-time tokenization for videos. In: Ranzato, M., Beygelzimer, A., Dauphin, Y., Liang, P., Vaughan, J.W. (eds.) Advances in Neural Information Processing Systems. vol.~34, pp. 12786--12797. Curran Associates, Inc. (2021), \url{https://proceedings.neurips.cc/paper_files/paper/2021/file/6a30e32e56fce5cf381895dfe6ca7b6f-Paper.pdf}

\bibitem{geminiteam2024geminifamilyhighlycapable}
Team, G., et~al.: Gemini: A family of highly capable multimodal models (2024), \url{https://arxiv.org/abs/2312.11805}

\bibitem{tulving1972episodic}
Tulving, E., et~al.: Episodic and semantic memory. Organization of memory  \textbf{1}(381-403), ~1 (1972)

\bibitem{wang2024internvideo2scalingfoundationmodels}
Wang, Y., Li, K., Li, X., Yu, J., He, Y., Wang, C., Chen, G., Pei, B., Yan, Z., Zheng, R., Xu, J., Wang, Z., Shi, Y., Jiang, T., Li, S., Zhang, H., Huang, Y., Qiao, Y., Wang, Y., Wang, L.: Internvideo2: Scaling foundation models for multimodal video understanding (2024), \url{https://arxiv.org/abs/2403.15377}

\bibitem{yang2025egolifeegocentriclifeassistant}
Yang, J., Liu, S., Guo, H., Dong, Y., Zhang, X., Zhang, S., Wang, P., Zhou, Z., Xie, B., Wang, Z., Ouyang, B., Lin, Z., Cominelli, M., Cai, Z., Zhang, Y., Zhang, P., Hong, F., Widmer, J., Gringoli, F., Yang, L., Li, B., Liu, Z.: Egolife: Towards egocentric life assistant (2025), \url{https://arxiv.org/abs/2503.03803}

\bibitem{zhang2025videollama3frontiermultimodal}
Zhang, B., Li, K., Cheng, Z., Hu, Z., Yuan, Y., Chen, G., Leng, S., Jiang, Y., Zhang, H., Li, X., Jin, P., Zhang, W., Wang, F., Bing, L., Zhao, D.: Videollama 3: Frontier multimodal foundation models for image and video understanding (2025), \url{https://arxiv.org/abs/2501.13106}

\bibitem{zhang2024how}
Zhang, M., Press, O., Merrill, W., Liu, A., Smith, N.A.: How language model hallucinations can snowball. In: Forty-first International Conference on Machine Learning (2024), \url{https://openreview.net/forum?id=FPlaQyAGHu}

\end{thebibliography}
\end{document}